\documentclass[letterpaper]{article} 
\usepackage{aaai24}  
\usepackage{times}  
\usepackage{helvet}  
\usepackage{courier}  
\usepackage[hyphens]{url}  
\usepackage{graphicx} 
\usepackage{natbib}  
\usepackage{caption} 
\usepackage{algorithm}
\usepackage{algorithmic}
\usepackage{enumitem}

\usepackage{multirow}
\usepackage{color}
\usepackage{amsmath,amssymb,amsthm}
\usepackage{mathtools}
\usepackage{subcaption}

\usepackage{newfloat}
\usepackage{listings}
\floatstyle{ruled}
\newfloat{listing}{tb}{lst}{}
\floatname{listing}{Listing}

\newcommand{\ie}{\textit{i}.\textit{e}.}
\newcommand{\eg}{\textit{e}.\textit{g}.}

\newcommand{\zj}[1]{{\color{black}#1}}

\newcommand{\hmy}[1]{{\color{black}#1}}
\title{PartSeg: Few-shot Part Segmentation via Part-aware Prompt Learning}
\author {
    Mengya Han\textsuperscript{\rm 1},
    Heliang Zheng\textsuperscript{\rm 2},
    Chaoyue Wang\textsuperscript{\rm 2},
    Yong Luo\textsuperscript{\rm 1},
    Han Hu\textsuperscript{\rm 3},
    Jing Zhang\textsuperscript{\rm 4}, \\
    Yonggang Wen\textsuperscript{\rm 5}
}
\affiliations {
    \textsuperscript{\rm 1}Wuhan University, China\\
    \textsuperscript{\rm 2}JD Explore Academy, China.
    \textsuperscript{\rm 3}Beijing Institute of Technology, China\\
    \textsuperscript{\rm 4}The University of Sydney, Australia\\
    \textsuperscript{\rm 5}Nanyang Technological University, Singapore\\
    
    myhan1996@whu.edu.cn, 
    zhenghllj@gmail.com,
    chaoyue.wang@outlook.com,
    luoyong@whu.edu.cn,
    hhu@bit.edu.cn,
    jingzhang.cv@gmail.com,
    ygwen@ntu.edu.sg 
}

\usepackage{bibentry}

\begin{document}

\maketitle

\begin{abstract}
 In this work, we address the task of few-shot part segmentation, which aims to segment the different parts of an unseen object using very few labeled examples. It is found that leveraging the textual space of a powerful pre-trained image-language model (such as CLIP) can be beneficial in learning visual features. Therefore, we develop a novel method termed PartSeg for few-shot part segmentation based on multimodal learning. Specifically, we design a part-aware prompt learning method to generate part-specific prompts that enable the CLIP model to better understand the concept of ``part'' and fully utilize its textual space. Furthermore, since the concept of the same part under different object categories is general, we establish relationships between these parts during the prompt learning process. We conduct extensive experiments on the PartImageNet and Pascal$\_$Part datasets, and the experimental results demonstrated that our proposed method achieves state-of-the-art performance. \zj{The source code and models will be made publicly available.}
\end{abstract}

\section{Introduction}~\label{introdcution}
\begin{figure}[t]
    \centering
    \includegraphics[width=\linewidth]{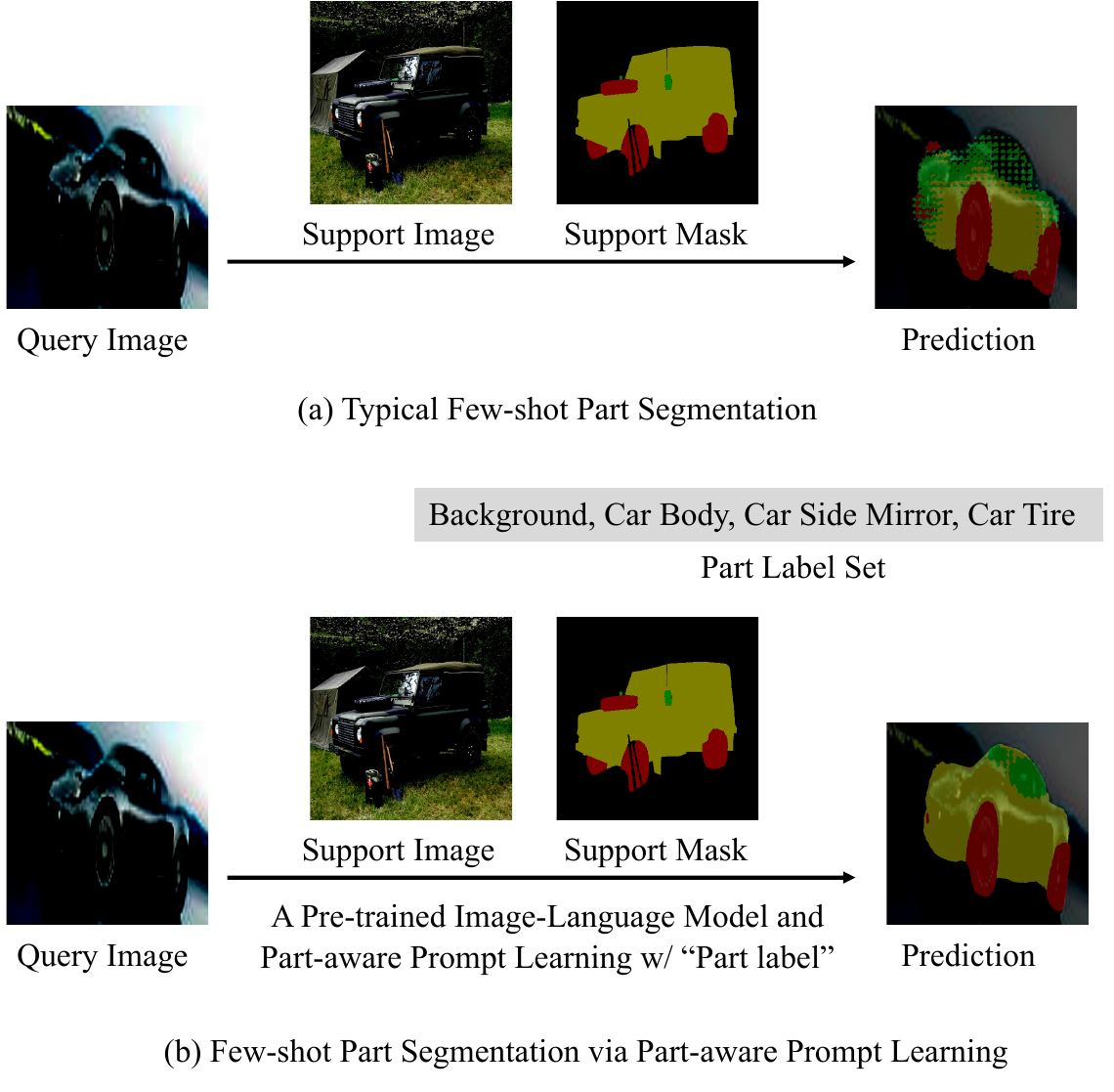}
    \caption{Comparison between (a) typical few-shot part segmentation approach and (b) the proposed few-shot part segmentation method via part-aware prompt learning. 
    }
    \label{fig:intro}
\end{figure}

In this work, we tackle the challenging task of few-shot part segmentation (FSPS), which segments different parts in an object of a novel category only seeing a few part-annotated examples.
We formulate FSPS as a guided segmentation task, where the guidance information is extracted from the annotated support set and then the query image is segmented using these extracted guidance information.
Figure~\ref{fig:intro}(a) illustrates the typical pipeline of few-shot part segmentation, where the query image is segmented based on the provided support images and their masks~\cite{zhang2021datasetgan,tritrong2021repurposing,2022han}.
However, such approaches may not be able to fully characterize an entire concept class by leveraging a limited number of annotated examples.
In contrast, humans learn new concepts efficiently by leveraging cross-modal information, such as textual (language) information~\cite{jackendoff1987beyond}. Such information embodies the way humans recognize objects, such as the concept of head or body which humans use to comprehend an object. By utilizing textual information pertaining to part concepts, we can enhance our comprehension of objects and their constituent parts.
Moreover, these part concepts often exhibit similarities across different object categories. Therefore, language information about parts is critical for enhancing generalization in few-shot tasks, enabling us to learn new concepts efficiently even with limited examples.

This motivates us to present PartSeg, a novel method to segment object parts in a query image using few-shot support images and their corresponding textual descriptions of the parts, as outlined in Figure~\ref{fig:intro}(b). \hmy{We are making the first attempt to leverage multi-modal learning for the few-shot part segmentation task}. Specifically, we leverage a pre-trained image-language model, CLIP~\cite{radford2021learning}, as the backbone to learn textual features from the textual descriptions of the parts. Our key insight is that the textual space of the image-language model can effectively guide the learning of visual features, enabling accurate recognition of object parts.
Moreover, the pre-trained CLIP model may not fully understand the concept of parts only through the textual descriptions. To reduce this semantic gap, we propose a part-aware prompt learning approach that utilizes both few-shot support images and their corresponding textual descriptions of the parts to generate textual features for better exploration of the textual space.

Furthermore, the concept of parts is general across different object categories. That is, the same parts in different categories should be proximal in the textual space. To achieve this, we employ the exponential moving average (EMA)~\cite{klinker2011exponential} algorithm to estimate the shared prompt tokens for the same parts. This \zj{simple yet effective} technique allows us to establish direct connections between them during the part-aware prompt learning process and leads to further performance enhancement.

Specifically, our PartSeg consists of three major components. The first component is the ``visual encoder''. In this component, we extract visual features for the support image and the query image based on the input images. Then, we obtain part-level visual prototypes by using masked average pooling to integrate the visual feature of the support image with its corresponding ground-truth mask. In the second component, the ``text encoder'', we utilize a text encoder from the pre-trained CLIP model to generate the textual feature from the input texts. The third component is our designed ``part-aware prompt learning'' module, which \zj{consists} of a part-specific prompt generator and a part-shared prompt estimator.
The part-specific prompt generator utilizes a network to produce part-specific tokens for each part class based on the learned part-level visual prototype. The part-shared prompt estimator leverages the EMA algorithm to estimate part-shared tokens for \zj{each same part} in different categories.
The part-specific tokens, the part-shared tokens, and \zj{the} textual description of parts (\zj{\ie}, part labels) are combined as the final prompt for the text encoder to generate the textual prototypes for each part. \hmy{The query image is segmented by comparing with part-level visual prototypes and textual prototypes. 
PartSeg enhances the capacity for generalization in few-shot tasks, enabling effective learning of unseen categories even with a limited number of examples.}

The main \zj{contribution of our method is two-fold}:
\begin{itemize}
    \item We propose a novel approach, PartSeg, for few-shot part segmentation. To the best of our knowledge, we are the first to leverage a pre-trained image-language model to guide the learning of visual features for few-shot part segmentation. Specifically, we utilize a text encoder to generate textual features for each part, which aids in more efficient learning of visual features.
    \item We develop a part-aware prompt learning module that \zj{consists} of a part-specific prompt generator and a part-shared prompt estimator, allowing us to more effectively leverage the textual space of the CLIP model.
\end{itemize}
We conduct extensive experiments on the PartImageNet~\cite{he2022partimagenet} and Pascal$\_$Part~\cite{chen2014detect} datasets, and \zj{the} results show that the proposed method \zj{achieves state-of-the-art (SOTA)} performance for few-shot part segmentation. For example, we achieve a $7.25\%$ relative improvement over the recent Lseg~\cite{lilanguage} counterpart in the cross-domain few-shot setting.

\section{Related Work}~\label{related_work}
\subsection{Few-shot Segmentation}~\label{few-shot_semantic_segmenattion}
Few-shot segmentation~\cite{wang2019panet,zhang2019canet,lang2022learning,zhuge2021deep,dong2021abpnet,liu2020dynamic,zhang2020sg,wu2021learning,zhang2021self,yang2020prototype,li2021adaptive} aims to segment objects in an image with only a few annotated examples. Many few-shot methods have been proposed by leveraging the meta-learning paradigm and prototype learning~\cite{finn2017model}.
Rather than studying few-shot segmentation at the instance level, this paper focuses on part-level few-shot segmentation. Specifically, we aim to address the challenge of recognizing and segmenting parts of objects with limited annotated data. 

\subsection{Few-shot Part segmentation}~\label{few-shot_part_segmentation}
Few-shot part segmentation is an extension of few-shot semantic segmentation that tackles the more challenging task of segmenting different parts within an object rather than just the entire object in an image.
In recent years, significant progress has been made in this field, which can be attributed to the advantages of generative models~\cite{tritrong2021repurposing,zhang2021datasetgan,baranchuk2021label,he2021progressive,2022han,he2023end}. However, existing few-shot part segmentation methods primarily rely on pre-trained generative models to segment \zj{a} single category (\eg, face), failing to segment the multiple categories due to the limitations of generative models~\cite{arjovsky2017wasserstein,arjovsky2017towards,karras2019style,dhariwal2021diffusion,ho2020denoising}. Therefore, once faced with unseen categories, it takes much time to train a generative model for \zj{a} specific category, which restricts the practical application in the open world.
In this work, we attempt to \zj{handle multiple categories on more complex datasets (\eg, PartImageNet~\cite{he2022partimagenet})}. To this end, we adopt a meta-leaning paradigm for the FSPS task and learn \zj{transferable knowledge across} categories. Once trained, the model can generalize well to unseen novel \zj{categories} using \zj{a few} annotated examples.

\subsection{Image-Language Models}~\label{image_language_models}
Several recent studies have demonstrated that image-language pre-training models can significantly boost the performance of visual tasks~\cite{radford2021learning,zhou2022learning,zhou2022conditional,zhang2023clamp,li2023referring}. For example, CLIP~\cite{radford2021learning} has exhibited strong zero-shot adaptation capabilities. CoOp~\cite{zhou2022learning} was developed to optimize a set of learnable prompt tokens as input for the text encoder. CoCoOp~\cite{zhou2022conditional}leverages visual information as a condition for generating image tokens as input for the text encoder.
Inspired by these approaches, we propose the PartSeg model, which is the first to \zj{leverage} a \zj{pre-trained} image-language model for few-shot part segmentation. Moreover, we introduce part-aware prompt learning to effectively explore the textual information of parts.

\section{Preliminaries}~\label{preliminaries}
\subsection{Problem Setting}~\label{problem_formulation}
Few-shot part segmentation aims to segment the different parts of an object in an image when only a few annotated images are available. In typical few-shot settings, the meta-learning paradigm (\ie, episodic training strategy)~\cite{vinyals2016matching,ravi2017optimization,snell2017prototypical} is adopted. Training and testing are conducted on two different datasets $\mathcal{D}_{base}$ and $\mathcal{D}_{novel}$. These datasets are disjoint with respect to object categories. $\mathcal{D}_{base}$ contains a set of base categories $\mathcal{C}_{base}$, each with sufficient annotated examples per category. Meanwhile, $\mathcal{D}_{novel}$ contains a set of novel categories $\mathcal{C}_{novel}$, each with only a few annotated examples per category. Our goal is to learn transferable knowledge on $\mathcal{D}_{base}$ and generalize it to $\mathcal{D}_{novel}$. Both datasets \zj{contain} numerous episodes (\ie, few-shot tasks), each of which contains a support set $\mathcal{S}$ and a query set $\mathcal{Q}$. During training, we iteratively sample an episode from $\mathcal{D}_{base}$ to optimize the model. The objective of the few-shot part segmentation task is to segment the query image based on the support set $\mathcal{S}$. Once the model is trained, we evaluate the performance on $\mathcal{D}_{novel}$ without any further optimization.


\begin{figure*}[t]
    \centering
    \includegraphics[width=1.0\textwidth]{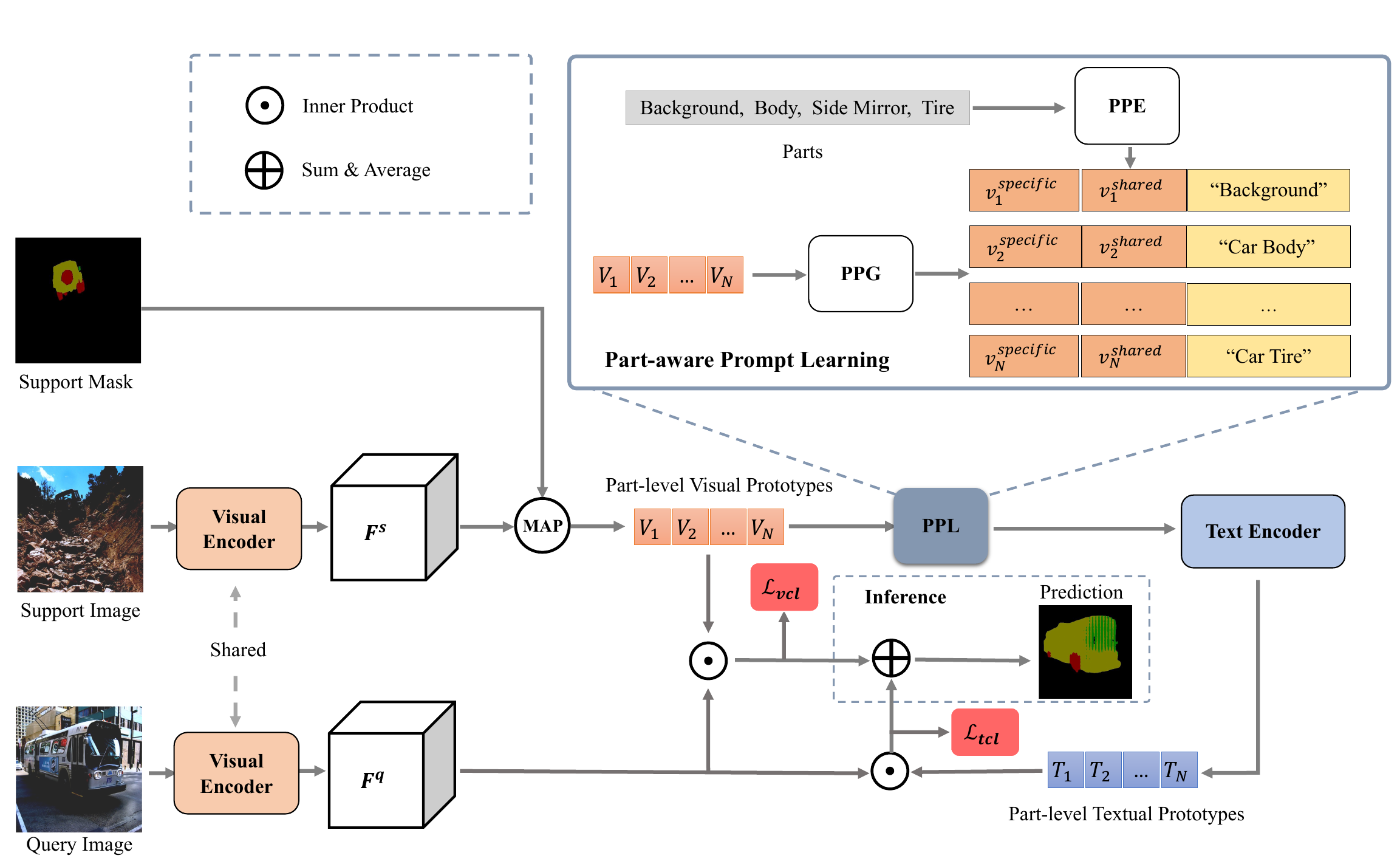}
    \caption{The framework of our PartSeg method for few-shot part segmentation. We first utilize a visual encoder to extract feature maps $F^{s}$ and $F^{q}$ from a given support image and query image, respectively. Then we apply masked average pooling (MAP) to generate the part-level visual prototypes $V_1, V_2, \cdots, V_N$ from the support feature $F^{s}$ and its support mask. To generate part-level textual prototypes $T_1, T_2, \cdots, T_N$, we propose the part-aware prompt learning (PPL) module that takes $V_1, V_2, \cdots, V_N$ and a part label set as \zj{a} condition. Specifically, our part-aware prompt learning (PPL) module \zj{consists} of a part-specific prompt generator (PPG) and a part-shared prompt estimator (PPE) to generate part-specific tokens $v^{specific}$ and part-shared tokens $v^{shared}$, respectively. These tokens are concatenated with the part label as the final prompt for text encoding. Finally, we utilize part-level visual prototypes and part-level textual prototypes to segment the query image. It is worth noting that the visual encoders for support and query images share the same weights in our method.}
    \label{fig:framework}
\end{figure*}

\subsection{ProtoNet}~\label{ProtoNet}
By leveraging the meta-learning paradigm~\cite{vinyals2016matching} and prototype learning approach~\cite{snell2017prototypical}, we have re-implemented ProtoNet as a visual baseline for few-shot part segmentation. As outlined in Figure~\ref{fig:framework}, it contains a visual encoder that extracts visual features for both the support and query images. We then employ a masked averaging pooling strategy to produce part-level visual prototypes. These prototypes are subsequently used to segment the query image via direct feature comparison.

\textbf{Visual Encoder.} We employ a pre-trained Vision-Transformer~\cite{dosovitskiy2020image,liu2021swin,ranftl2021vision} as the visual encoder to learn visual features for both the support and query images. The output of the visual encoder is defined as $F \in \mathbb{R}^{C \times H \times W}$, where $C$, $W$, and $H$ denote the number of channels, the width, and the height of the output, respectively. We refer to the feature vector of pixel $(i,j)$ as $F_{ij} \in \mathbb{R}_{C}$.
As illustrated in Figure~\ref{fig:framework}, in a few-shot part segmentation task, we have a support image and its corresponding segmentation mask, along with an unlabeled query image from the same object category. We assume that this object category has $N$ part labels, and the objective is to segment the query image based on the support image and its segmentation mask.  Specifically, we first embed the support image and the query image into support feature $F^{s} \in \mathbb{R}^{C \times H \times W}$ and query feature $F^{q} \in \mathbb{R}^{C \times H \times W}$, respectively. Then, we adopt a masked averaging pooling strategy to obtain a comprehensive representation of each specific part class, \zj{\ie, a} part-level visual prototype.
The part-level visual prototype $V_k$ of the $k$th part class in the support image is computed as:
\begin{equation}
    V_k = \frac{ {\textstyle \sum_{i,j}^{H,W}F^{s}_{ij}\mathbb{I}\left [ M^{s}_{ij} = k\right ]} }{ {\textstyle \sum_{i,j}^{H,W}\mathbb{I}\left [ M_{ij} = k\right ]} },
\end{equation}
where $M^{s}_{ij} \in \left \{ 1,...,N \right \} $ \zj{represents} the ground-truth label of pixel $(i,j)$. $\mathbb{I}$ is an indicator function that outputs a value of 1 if the argument is true, and 0 otherwise.

\textbf{Visual Contrast Loss.} 
After learning the part-level visual prototypes and the query feature $F^{q} \in \mathbb{R}^{C \times H \times W}$, we proceed to correlate them using inner product and generate a correlation tensor of size $H \times W \times N$, where each element is defined as $f^{v}_{ijk} = F^{q}_{ij} \cdot V_k$.
During training, our aim is to encourage the visual encoder to produce pixel features that are similar to the part-level visual prototype of the corresponding ground-truth part class. To this end, we maximize the inner product of $f^{v}_{ijk}$ that corresponds to the ground-truth part label $k$ of pixel $(i,j)$. We then compute the predicted probability that pixel $(i,j)$ belongs to part class $k$ as follows: 
\begin{equation}~\label{eq:p}
p^{v}_{ijk}= \frac{exp(f^{v}_{ijk})}{ {\textstyle \sum_{n=1}^{N}} exp(f^{v}_{ijn})}.
\end{equation}
In this work, we employ cross-entropy loss as the segmentation loss, which is defined as:
\begin{equation}
    L_{vcl} =  -\sum_{i,j=1}^{H,W} \sum_{k=1}^{N} -\mathbb{I}[M^{q}_{i,j}=k] log(p^{v}_{ijk}),
\end{equation}
where $M^{q}$ is the ground truth segmentation mask of the query image.

\section{Methodology}
\subsection{Overview}~\label{overview}
In this work, we build upon the visual baseline ProtoNet and develop a novel few-shot part segmentation approach called PartSeg by leveraging a pre-trained image-language model. Figure~\ref{fig:framework} illustrates the overall framework of the proposed PartSeg, which comprises three major components: a visual encoder, part-aware prompt learning (PPL), and a text encoder. For a given few-shot part segmentation task, we \zj{first} utilize the visual encoder to learn visual features for support and query images, and obtain part-level visual prototypes from support images and their corresponding ground-truth segmentation masks via masked average pooling. Subsequently, we design a part-specific prompt generator (PPG) and a part-shared prompt estimator to generate part-specific tokens and part-shared tokens for each part using the part-level visual prototypes and their textual description of part classes (\eg, part labels). These tokens are then concatenated with textual tokens of the part class learned from the pre-trained language model tokenizer to form a final prompt that is fed into the text encoder to produce part-level textual prototypes.
Finally, we segment the query image by comparing its feature with both the part-level visual prototypes and part-level textual prototypes. The details of \zj{the} visual encoder \zj{have been} outlined in Section~\ref{ProtoNet}. To provide a more comprehensive understanding of this process, we offer a detailed description of \zj{each component} in the following parts.

\subsection{Part-aware Prompt Learning}~\label{part_aware_prompt_learning}
The pre-trained CLIP \zj{model} has demonstrated impressive performance on computer vision tasks. However, these models rely on textual descriptions of objects for training and may struggle to fully comprehend the concept of a part through text alone, such as understanding the ``body'' of a car just from its textual description. To address this issue, we propose a PPL approach that leverages few-shot support images and corresponding text descriptions of parts to more effectively explore the textual space of the CLIP model. Instead of developing global tokens for all part classes, we introduce both part-specific and part-shared tokens to generate more informative part-level textual prototypes. Specifically, our PPL includes a part-specific prompt generator that produces part-specific tokens for each part. Additionally, since the concept of parts is general, we aim to establish connections between \zj{the} same parts in different categories. To achieve this, we design a part-shared prompt estimator (PPE), which generates part-shared tokens using the EMA algorithm during the prompt learning process. We then concatenate the resulting part-specific and part-shared tokens to form visual tokens. An overview of our proposed PPL approach is presented in Figure~\ref{fig:framework}. In the following, we provide a detailed description of the proposed part-specific prompt generator and part-shared prompt estimator.

\textbf{Part-specific Prompt Generator}.
To effectively explore semantic information from part labels, we use visual information from a support image as a condition to generate part prompt tokens that aid in understanding the concept of parts. Specifically, we train a shallow neural network to learn part-specific tokens for each part based on their corresponding visual prototypes. First, we obtain the support feature representation $F^{s}$ from a visual encoder given the input image $I^{s}$. Then, we apply \zj{the} mask average pooling strategy to the support feature and the corresponding ground-truth mask to obtain the part-level visual prototypes $V_1, V_2, ..., V_N$. Finally, we use the shallow neural network $f$ to generate the part-specific tokens $ v^{specific}_k$ for part class $k$ as follows:
\begin{equation}
    v^{specific}_k = f(V_k), k=1,2, ..., N.
\end{equation}
The learned part-specific tokens are helpful for distinguishing different parts of an object.

\textbf{Part-shared Prompt Estimator}. Since the concept of a part is general, parts that are the same across different objects (the ``body'' of the Bicycle and ``body'' of the Car) should be closer together in the textual space. To better leverage these shared concepts and enhance generalizability, we propose a part-shared prompt estimator, which establishes a direct relationship between the same parts under different objects. We leverage the EMA algorithm to learn part-shared tokens for each part. The process begins with the random initialization of learnable feature vectors as both the current part-shared tokens and part-shared tokens for each part. We subsequently incorporate the current part-shared tokens of part class $k$ with part-shared tokens. Specifically, part-shared tokens $v^{shared}_k$ are updated iteratively using the EMA algorithm during training:
\begin{equation}
    v^{shared}_k \gets m v^{shared}_k + (1-m) v^{cur}_k,
\end{equation}
where $v^{cur}_k$ \zj{represents} the current part-shared tokens of part class $k$, \zj{and} $m$ is the momentum coefficient. By learning part-shared tokens from the same parts of different objects, the model can better understand the commonalities between those parts and transfer knowledge from one object to another based on shared parts.

Part-shared tokens $v^{shared}_k$ and part-specific tokens $v^{specific}_k$ are concatenated to form the visual tokens of part class $k$. The visual tokens can be formulated as follows:
\begin{equation}
    v_k = [v^{specific}_k][v^{shared}_k].
\end{equation}

\subsection{Text Encoder}
The text encoder embeds a set of $N$ part labels into a feature space, generating $N$ feature vectors $T_1$, $T_2$, ..., $T_N$, where $T \in \mathbb{R}_{C}$. In this work, we use the text encoder from the CLIP model. To obtain the textual tokens of each part class $t_k = \psi (k)$, we use a text tokenizer from the pre-trained image-language model to encode the textual description of the part. Then, the textual tokens are concatenated with the visual tokens, resulting in a final prompt: $prompt_k = [v_k][t_k]$. We leverage the CLIP model's text encoder $T()$ to produce the part-level textual prototype $T_k$ from both visual (support images) and textual (part class names) information for part class $k$. The output is denoted as follows:
\begin{equation}
    T_k = T(prompt_k) \in \mathbb{R}^{C}.
\end{equation}

\subsection{Training objective}
\textbf{Textual Contrast Loss.}
 To calculate the textual contrast loss, we compute the inner product between the part-level textual prototypes and the query feature, resulting in a correlation tensor of size $H \times W \times N$. Each element $f^{t}_{ijk} = F^{q}_{ij} \cdot T_k$ of this tensor represents the similarity between the query feature of pixel $(i,j)$ and the part-level textual prototype $T_k \in \mathbb{R}^{C}$ for the part label class $k$. 
\hmy{Subsequently, we refer to Eq.~\eqref{eq:p} to calculate the predicted probability $p^{t}_{ijk}$, indicating the likelihood of the pixel $(i,j)$ belonging to part class $k$.
The textual contrast loss is defined} as follows:
\begin{equation}
    L_{tcl} =  -\sum_{i,j=1}^{H,W} \sum_{n=1}^{N} - y_{ij} log(p^{t}_{ijk}).
\end{equation}

\textbf{Training Loss.} To train our PartSeg model, the total loss is the combination of visual contrast loss and textual contrast loss, \zj{which is defined} as follows:
\begin{equation}
    L = L_{vcl} + L_{tcl}.
\end{equation}

\begin{figure*}[!t]
  \centering
 \includegraphics[width=1\linewidth]{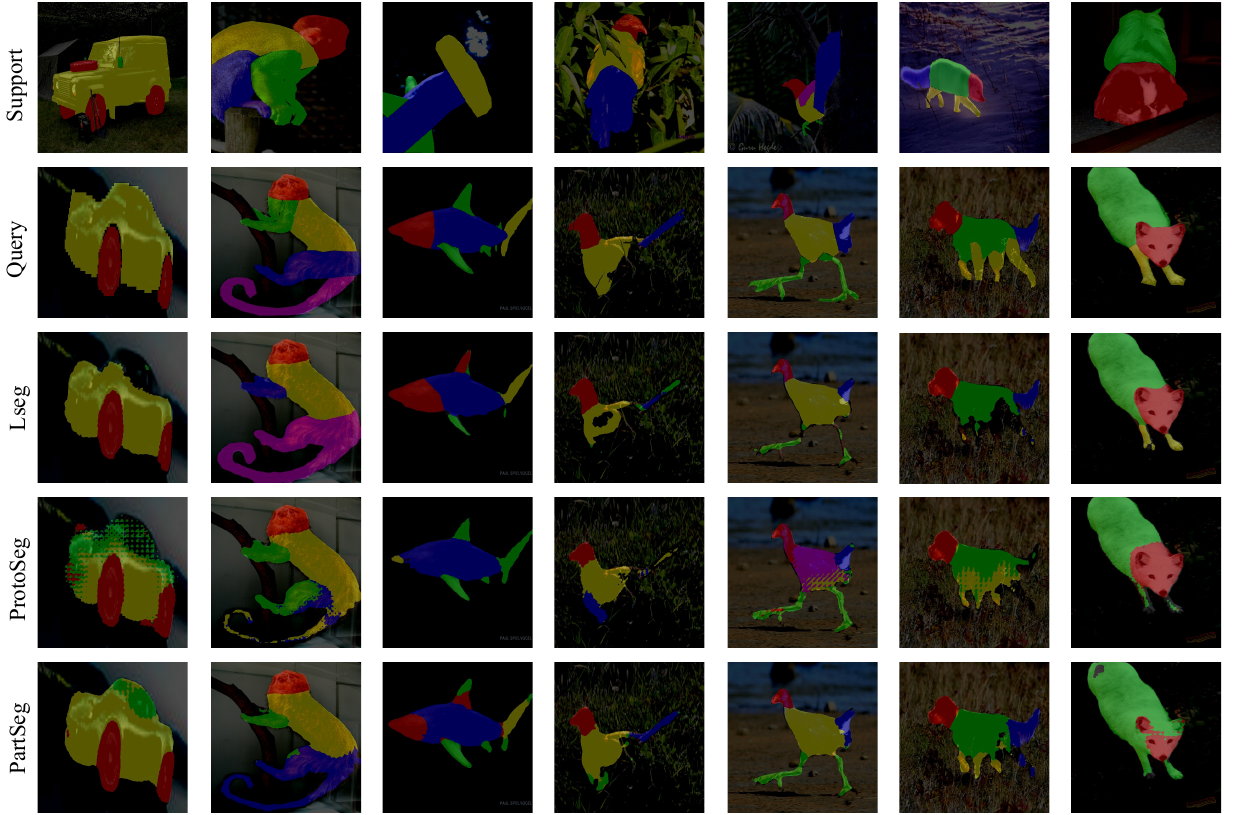}
  \caption{Qualitative comparison on the PartImageNet dataset. The last column shows a failure case.}
  \label{fig:result}
\end{figure*}

\begin{table*}[!t]
{
\small
\caption{Few-shot part segmentation performance (mIoU) on the PartImageNet dataset. ProtoNet is the our re-implemented visual baseline.}
\label{tab:partimagenet}
\begin{center}
\begin{tabular}{ c | c | c |c c cccc}
\hline
Method & Backbone & Text Encoder& Novel Set 1 & Novel Set 2 & Novel Set 3 & Novel Set 4 \\
\hline \hline
Lseg & -& ViT-B/32 & 0.5206 & 0.6099 & 0.5708 & 0.6131\\
ProtoNet &ViT-L/16 & -& 0.5233 & 0.6076  & 0.5649 & 0.5799\\
PartSeg &ViT-L/16 & ViT-B/32& 0.5434 &  0.6250 & 0.5917 & 0.6340\\
    \hline
    \end{tabular}
\end{center}
}
\end{table*}

\section{Experiments}
\subsection{Experiment setup}


\textbf{Datasets}.
We evaluated our model on the \zj{challenging} PartImageNet~\cite{he2022partimagenet} and Pascal$\_$Part~\cite{chen2014detect} datasets.

\begin{itemize}
\item PartImageNet is a subset of ILSVRC-12~\cite{imagenet} that includes part-level annotations. This dataset consists of 11 super-categories.
Since the \zj{part-annotated} datasets are rare, we randomly divided the PartImageNet dataset to construct four splits, each containing 9 super-categories for training and 2 super-categories for testing. 
\item Pascal$\_$Part is a part-annotated dataset that provides additional annotations for~\cite{everingham2010pascal}. The dataset includes segmentation masks for each part of the object and contains 20 categories. However, some categories do not have a consistent set of parts (\eg, boat), making them unsuitable for part segmentation. Therefore, we have excluded these categories from this dataset to perform few-shot part segmentation tasks.
\end{itemize}

\textbf{Implementation \zj{D}etails}. We adopted the episodic training strategy commonly used in previous few-shot works~\cite{vinyals2016matching,snell2017prototypical}. Specifically, we randomly selected pairs of images from the training set where each pair contained two images sharing a common category label (\eg, Car or Biped). One image served as the support image and was fed into the network along with its corresponding grounding truth mask, while another image was treated as the query image and used its grounding truth mask to compute the loss. We leveraged a ViT-L/16 backbone as the visual encoder and adopted the text encoder provided by CLIP-ViT-B/32. We initialized the visual encoder's backbone with official pre-trained weights from ViT on ImageNet. During training, we froze the text encoder and only updated the weights of other modules. We trained \zj{the model} using \zj{the} SGD \zj{optimizer} with a momentum of 0.9 and \zj{the} polynomial learning rate scheduler with a decay rate of 0.9. The batch size was set to 1, and we run the training on 4 NVIDIA 3090RTX GPUs.

\textbf{Baselines}. 
\hmy{In this work, we introduce a new few-shot part segmentation task setting to handle multiple categories effectively based on the meta-learning paradigm. To this end, we propose a novel multi-modal approach, leveraging a few annotated examples and language information about parts. Moreover, it's noteworthy that there exists a notable absence of comparative methodologies within this new task setting.} As a result, we conducted a comparison between our approach and two baseline methods that leverage different modalities: a text-based part segmentation approach and a visual-based part segmentation approach. 
\begin{itemize}
\item Lseg~\cite{lilanguage} — this method is a text-based part segmentation method that exploits the given part label set to segment the unlabeled query image.
\item ProtoNet - this method is a visual-based part segmentation approach that utilizes few-shot support images to segment unlabeled query images. ProtoNet is our re-implemented visual baseline method based on the prototype learning paradigm~\cite{snell2017prototypical} commonly used in few-shot learning.
\end{itemize}

\begin{table*}[t]
{
\small
\caption{Ablation study of different components \zj{in PartSeg} on the PartImageNet dataset.}
\label{tab:components}
\begin{center}
\begin{tabular}{ c c|c c  c c}
\hline
 Text Encoder & PPL &Novel Set 1 & Novel Set 2 & Novel Set 3 & Novel Set 4\\
\hline \hline
&  & 0.5233 & 0.6076  & 0.5649 & 0.5799\\
\checkmark  & & 0.5380 & 0.6198 & 0.5851 & 0.6314 \\
 \checkmark & \checkmark & 0.5434 &  0.6250 & 0.5917 & 0.6340\\
    \hline
    \end{tabular}
\end{center}
}
\end{table*}

\begin{table*}[t]
{
\small
\caption{Ablation study of the prompt learning design on the PartImageNet dataset.}
\label{tab:prompt}
\begin{center}
\begin{tabular}{ c| c c  c c}
\hline
\zj{Prompt} Design & Novel Set 1 & Novel Set 2 & Novel Set 3& Novel Set 4 \\
\hline \hline
LGP & 0.5337&  0.6131 & 0.5627 & 0.6163\\
LPP & 0.5396  &  0.6201 & 0.5808 & 0.6238\\
Our PPL & 0.5434 &  0.6250 & 0.5917 & 0.6340 \\
    \hline
    \end{tabular}
\end{center}
}
\end{table*}

\begin{table}[!t]
{
\small
\caption{Cross-domain few-shot part segmentation in \zj{terms} of mIoU.}
\label{tab:cross-domain}
\begin{center}
\begin{tabular}{ c | c}
\hline
Method & PartImageNet $\to$  Pascal$\_$Part\\
\hline \hline
Lseg & 0.4579\\
ProtoNet &0.4830\\
PartSeg & 0.4911\\
    \hline
    \end{tabular}
\end{center}
}
\end{table}

\begin{table}[!t]
{
\small
\caption{Impact of hyper-parameter $m$ in EMA.}
\label{tab:ema}
\begin{center}
\begin{tabular}{ c | c}
\hline
$m$ & Novel Set 1\\
\hline \hline
0 & 0.5334 \\
0.5 & 0.5385  \\
0.9 & 0.5413\\
0.99 & 0.5434 \\
    \hline
    \end{tabular}
\end{center}
}
\end{table}

\textbf{Evaluation Metric}. The evaluation metric employed in this work was the mean intersection-over-union (mIoU), which is commonly used for semantic segmentation tasks. The mIoU measures the overlap between predicted and ground truth segmentation maps and provides a reliable assessment of segmentation accuracy.

\subsection{Main Results}
We conducted experiments on the PartImageNet dataset to evaluate the proposed PartSeg. We compared \zj{it} with representative methods using the mIoU metric, and the results are presented in Table~\ref{tab:partimagenet}. We achieved superior performance across all splits, even outperforming the strongest baselines by a significant margin. Specifically, we obtained the highest score of 0.5434 mIoU on the Novel Set 1 dataset, surpassing Lseg and ProtoNet by 4.3\% and 3.8\%, respectively. Additionally, our approach demonstrated superior performance on the Novel Set 2, Novel Set 3, and Novel Set 4 datasets, where we achieved 2.4\% and 2.8\% relative improvement over the Lseg and ProtoNet on the Novel Set 2 dataset, 3.6\% and 4.7\% for Lseg and ProtoNet on the Novel Set 3 dataset, and 3.4\% and 9.3\% for Lseg and ProtoNet on the Novel Set 4 dataset. These substantial improvements can be attributed to the leverage of a pre-trained image-language model and the designed part-aware prompt learning strategy. Furthermore, we provide qualitative comparisons in Figure~\ref{fig:result}.

\subsection{Ablation studies}

\textbf{Effectiveness of Different Components}. 
Our approach consists of two new components: a text encoder and a part-aware prompt learning module. To evaluate their effectiveness, we conducted comprehensive ablation studies on the PartImageNet dataset by removing these components from our PartSeg model to obtain the baseline model (\ie, ProtoNet), which only uses the few-shot support image to segment the query image. The results of our ablation analysis are presented in Table \ref{tab:components}. Our findings indicate that utilizing the text encoder resulted in significant performance improvements of up to 5.15\% compared to the baseline model. Moreover, employing a part-aware prompt learning strategy to generate the prompt for extracting part-level textual prototypes resulted in further enhancements in performance. With all components, our method achieves the best performance and surpasses the state-of-the-art. These results demonstrate the effectiveness of our proposed text encoder and part-aware prompt learning module.

\textbf{Part-aware Prompt Learning}.
To \zj{investigate the design choice} of prompt learning and \zj{validate the superiority of the proposed PPL over other designs}, we compared it to two alternative methods: (i) Learnable Global Prompt (LGP), which learns common tokens for all parts from the image's visual feature, and (ii) Learnable Part-specific Prompt (LPP), which develops part-specific tokens for each part using the corresponding part-level visual prototype. From Table~\ref{tab:prompt}, we observed that (1) developing part-specific tokens for each part is more effective than using common tokens for all parts, and (2) estimating part-shared tokens for the same parts improves generalizability across categories since the concept of parts is general.

\textbf{PartImageNet $\to$ Pascal$\_$Part}.
In this paper, we present a \zj{new} cross-domain \zj{generalization} setting~\cite{boudiaf2021few} for few-shot part segmentation, where we train models on the split of the PartImageNet dataset and evaluate them \zj{directly} on the split of the Pascal$\_$Part dataset. The comparison results presented in Table~\ref{tab:cross-domain} demonstrate that our proposed PartSeg approach outperforms \zj{the baseline} methods. The cross-domain evaluations further demonstrate the effectiveness of our method.

\textbf{Hyper-parameter $m$}.
To estimate the shared tokens of the same parts from different categories during the part-aware prompt learning process, we utilize the EMA algorithm. In Table~\ref{tab:ema}, we report the impact of the hyper-parameter $m$ in the EMA. We observe that setting $m$=0.99 yields the best performance, but other values also lead to favorable results.

\section{Conclusion}
In this paper, we present PartSeg, \zj{the first} multi-modal approach for few-shot part segmentation. Our proposed method effectively segments the unlabeled query image using a few support images and a pre-trained image-language model. Moreover, we developed a part-aware prompt learning module that enables our approach to benefit from the textual space of the CLIP model and facilitates understanding the concept of ``part''. Extensive experiments \zj{on challenging datasets} demonstrate the effectiveness of \zj{the proposed PartSeg method}. In the future, we intend to improve the performance of \zj{our PartSeg by leveraging large-scale unlabeled data in either the semi-supervised learning paradigm or the pre-training and fine-tuning paradigm and applying it to the video domain as well as other downstream tasks such as image generation and editing}.



\end{document}